 \documentclass[smallabstract,smallcaptions]{dccpaper}

\usepackage{epsfig}
\usepackage{amssymb,amsmath,amsthm,enumitem}
\usepackage{color}
\usepackage{url}
\usepackage{epsfig}
\usepackage{color}
\usepackage{url}
\usepackage{enumitem}
\usepackage{graphicx}
\usepackage{algorithm}
\usepackage{algorithmicx}
\usepackage[noend]{algpseudocode}
\algnewcommand{\LineComment}[1]{\State \(//\) #1}

\usepackage{titlesec}
\titlespacing*{\section}{0pt}{1\baselineskip}{0.4\baselineskip}
\titlespacing*{\subsection}{0pt}{0.5\baselineskip}{0.2\baselineskip}
\titlespacing*{\subsubsection}{0pt}{0.4\baselineskip}{0.4\baselineskip}
\titlespacing*{\paragraph}{0pt}{0.4\baselineskip}{0.4\baselineskip}

\newlength{\figurewidth}
\newlength{\smallfigurewidth}

\setlist[itemize]{noitemsep, topsep=1pt}
\usepackage[utf8]{inputenc}

\begin{document}

\title{\bf Learned Neural Iterative Decoding for \\Lossy Image Compression Systems}
% Old title: Learned Iterative Decoding for Lossy Image Compression Systems
%Iterative Refinement of Quantized\\ Compact Representations for Lossy Image Compression Systems
\author{Alexander G. Ororbia\textsuperscript{$\star$}, 
Ankur Mali\textsuperscript{\textdagger}, 
Jian Wu\textsuperscript{\textdaggerdbl},\\  
Scott O'Connell\textsuperscript{\textdagger}, 
William Dreese\textsuperscript{\textdagger}, 
David Miller\textsuperscript{\textdagger}, 
C. Lee Giles\textsuperscript{\textdagger}\\ [0.5em] 
\textsuperscript{$\star$ Rochester Institute of Technology, Rochester, NY, 14623, USA}\\
\textsuperscript{\textdagger\ Pennsylvania State University, University Park, PA, 16802, USA}\\
\textsuperscript{\textdaggerdbl\ Old Dominion University, Norfolk, VA, 23529, USA}
%%{\small\begin{minipage}{\linewidth}\begin{center}
%%\begin{tabular}{ccc}
%%\hspace{-0.15in}  %AO: shifted left in the world of "negative spaces"
%%\textsuperscript{\star}Rochester Institute of Technology && \textsuperscript{\textdagger}Pennsylvania State University \\
%%20 Lomb Memorial Dr. && E397 Westgate Building \\
%%Rochester, NY, 14623 && University Park, PA, 16802\\ 
%%United States && United States \\
%\url{ago@cs.rit.edu} && \url{aam35@psu.edu}
%%\end{tabular}
%%\end{center}\end{minipage}}
%\thanks{* Only emails for corresponding authors have been provided.}
%%\thanks{This work has been submitted to the IEEE for possible publication. Copyright may be transferred without notice, after which this version may no longer be accessible.}
}
\maketitle
\thispagestyle{empty}
\setcounter{page}{1}
\pagenumbering{roman}

\begin{abstract}
For lossy image compression systems, we develop an algorithm, \emph{iterative refinement}, to improve the decoder's reconstruction compared to standard decoding techniques. Specifically, we propose a recurrent neural network approach for nonlinear, iterative decoding. Our decoder, which works with any encoder, employs self-connected memory units that make use of causal and non-causal spatial context information to progressively reduce reconstruction error over a fixed number of steps. We experiment with variants of our estimator and find that iterative refinement consistently creates lower distortion images of higher perceptual quality compared to other approaches. Specifically, on the Kodak Lossless True Color Image Suite, we observe as much as a $0.871$ decibel (dB) gain over JPEG, a $1.095$ dB gain over JPEG 2000, and a $0.971$ dB gain over a competitive neural model.
\end{abstract}

\Section{Introduction}
\label{intro}
Image compression is a problem that has been at the core of signal processing research for decades. 
Recent successes in the application of deep neural networks (DNNs) to problems in speech processing,  computer vision, and natural language processing have sparked the development of neural-based approaches to this challenging problem. 
%\cite{hinton2012deep}
%\cite{krizhevsky2012imagenet}
%\cite{mikolov2010recurrent}
However, most efforts strive to design end-to-end neural-based systems, which require designing and training effective encoding and decoding functions as well as quantizers.
%, the process of mapping a large set of values, such as the set of real-valued numbers, to a smaller and discrete set, such as the set of integers. 
This poses a particular challenge for back-propagation-based learning since the quantizer is discrete and thus not differentiable. 
In order to learn encoders, most work generally attempts to formulate relaxations or differentiable (``soft'') approximations to quantization \cite{agustsson2017}. % e.g. soft entropy coding,
Furthermore, most image compression systems decode spatial blocks separately, without considering the spatial dependencies with surrounding blocks. Our proposed algorithm takes a novel, neural network decoding approach that exploits spatially non-causal statistical dependencies, with the potential for achieving improved decoding accuracy 
%at a given bitrate 
given an encoded bit stream. 
%We propose an RNN decoder to fully exploit these statistical dependencies.

%In this paper, we take a different and less complicated tack to avoid the difficulties that arise with quantization.
Instead of building an end-to-end system, we seek to answer the question: given any encoder and quantization scheme what would a optimal non-linear decoder look like? Focusing on the decoder of the system allows us to side-step issues like approximating quantization and to take advantage of well-developed encoders and quantization operations (i.e., as in JPEG/JPEG 2000) while still giving us the opportunity to reduce distortion through improved decoder optimization.
To this end, we propose a novel \emph{non-linear} estimator as an image decoder. While our approach is general and can handle any encoder, here we focus on JPEG-encoded bit-streams.
While the local memory of recurrent neural networks (RNNs) is typically used to capture short-term causal context, we re-purpose this memory to gradually improve its reconstruction of image patches, an approach we call \emph{iterative refinement}.

\section{Related Work}
\label{related_work}
Traditional lossy compression techniques for still images exploit the fact that most of the image energy is concentrated in low spatial frequencies, with sparse content found in high frequencies. Thus, strategies that combine transform coding with optimized bit allocation are used to give good compression performance at acceptable computational and memory complexity as compared to high-dimensional vector quantization. JPEG employs an $8\times8$ block discrete cosine transform (DCT), followed by run-length coding that exploits the sparsity pattern of the resultant frequency coefficients \cite{wallace1991jpegstandard}. % \cite{takamura1994coding}.
JPEG 2000 (JP2) leverages a multi-scale discrete wavelet transformation (DWT) and applies a uniform quantizer followed by an adaptive binary arithmetic coder and bit-stream organization \cite{rabbani2002jpeg2000}. %%The key components underlying these compression approaches, transformation, quantization, coding, are often manually tuned/adjusted.
%They both contain three components: transformer, quantizer, and entropy coder, which are separately optimized, often through manual adjustment of parameters. 

Although widely used, DCT and DWT are ``universal'', i.e., they do not exploit the particular pixel distribution of the input images. Statistical learning techniques, in contrast, have greater power to represent non-linear feature combinations and to automatically capture latent statistical structure. %%The last decade has witnessed a proliferation of image compression methods using (deep) neural networks.
%, e.g., \cite{aiazzi2002context,baig2017learning,santurkar2017generative,johnston2017improved}.
%agustsson2017,
%aiazzi2000icip,
For instance, using a sample of the Outdoor MIT Places dataset, JPEG 2000 and JPEG achieve 30 and 29 times compression, respectively, but a neural-based technique achieves a better peak signal-to-noise ratio (PSNR) at just a quarter of the bitrate of the JPEG and JPEG 2000 with comparable visual quality \cite{toderici2016full}. Van den Oord et al. presented a neural network that sequentially predicts image pixels along two spatial dimensions \cite{oord2016pixel}. 
%%Architectural innovations in deep recurrent networks, including fast two-dimensional recurrent layers and an effective use of residual connections, have yielded log-likelihood scores on natural images that are considerably better than prior state-of-the-art \cite{oord2016pixel}.

%Early machine learning approaches made use of K-means clustering to separate foreground content from background content \cite{bottou1998djvu}. Autoencoders have also been used to reduce the dimensionality of images \cite{hinton2006science}. 
In \cite{gregor2016conceptualcompression}, a representation learning framework using the variational autoencoder for image compression was proposed. Toderici et al. (2015) proposed a framework for variable-rate image compression and an architecture based on convolutional and deconvolutional LSTM models \cite{toderici2015}. Later on, Toderici et al. (2016) proposed several architectures consisting of an RNN-based encoder and decoder, a binarizer, and an ANN for encoding binary representations. This was claimed to be the first ANN architecture that outperformed JPEG at image compression across most bitrates on the Kodak image dataset \cite{toderici2016full}. Other methods have been proposed using autoencoders, generative adversarial networks, etc. \cite{Theis2017,rippel2017real}. % or even CNNs \cite{jiang2017end} 
%More recently, using generative adversarial network (GAN), Rippel et al. (2017) built a multiscale adversarial training model, resulting in visually pleasing images. The evaluation indicates that this algorithm surpasses all commercial image compression techniques, and moreover run in tens of milliseconds for RAISE-1k $512\times768$ dataset with high MS-SSIM \cite{wang2004imagequality}. 
%In addition, Theis et al. proposed a compression autoencoder that achieves continuous relaxation using a smooth approximation of the discrete rounding function and a discrete entropy loss upper bound \cite{Theis2017}.
%Balle et al. used generalized divisive normalization (GDN) for joint nonlinearity and replaced rounding quantization with additive uniform noise for continuous relaxation \cite{Balle2016}.
%Li et al. proposed a weighted content compression method based on image importance maps \cite{Li2017}. For lossless compression, the method proposed by Theis et al. and van den Oord et al. achieve promising results.
%Jiang et al. proposed a CNN-based compression framework to achieve high-quality image compression at low bit rates \cite{jiang2017end}. 

%One limitation of previous work is that 
Most prior work designs encoders and thus quantization schemes, e.g., \cite{toderici2015}. Our method is different from these in two ways: 1) we propose an iterative, RNN-based estimator for decoding instead of using transformations, 2) our algorithm introduces a way to effectively exploit both causal and non-causal 
%(neighborhood of surrounding patches) 
information to improve low bitrate reconstruction. Our model applies to any image decoding problem and can handle a wide range of bitrate values.

\section{A Nonlinear Estimator for Iterative Decoding}
\label{iterative_refinement}
% Motivate reason why a neural compression system is a good idea...
 % and assumes that the images has been encoded within a specified target bitrate.
%Since our aim is to reconstruct test images not seen by the learner before in training, there is no need to consider the compression rate associated with our decoder's program and we do not need to consider any ``side information'' associated with the decoder's parameters. This is further bolstered by the fact that our bitrate is determined by the encoder such as JPEG.

\begin{figure}[t]
\centering
\includegraphics[width=2.75cm]{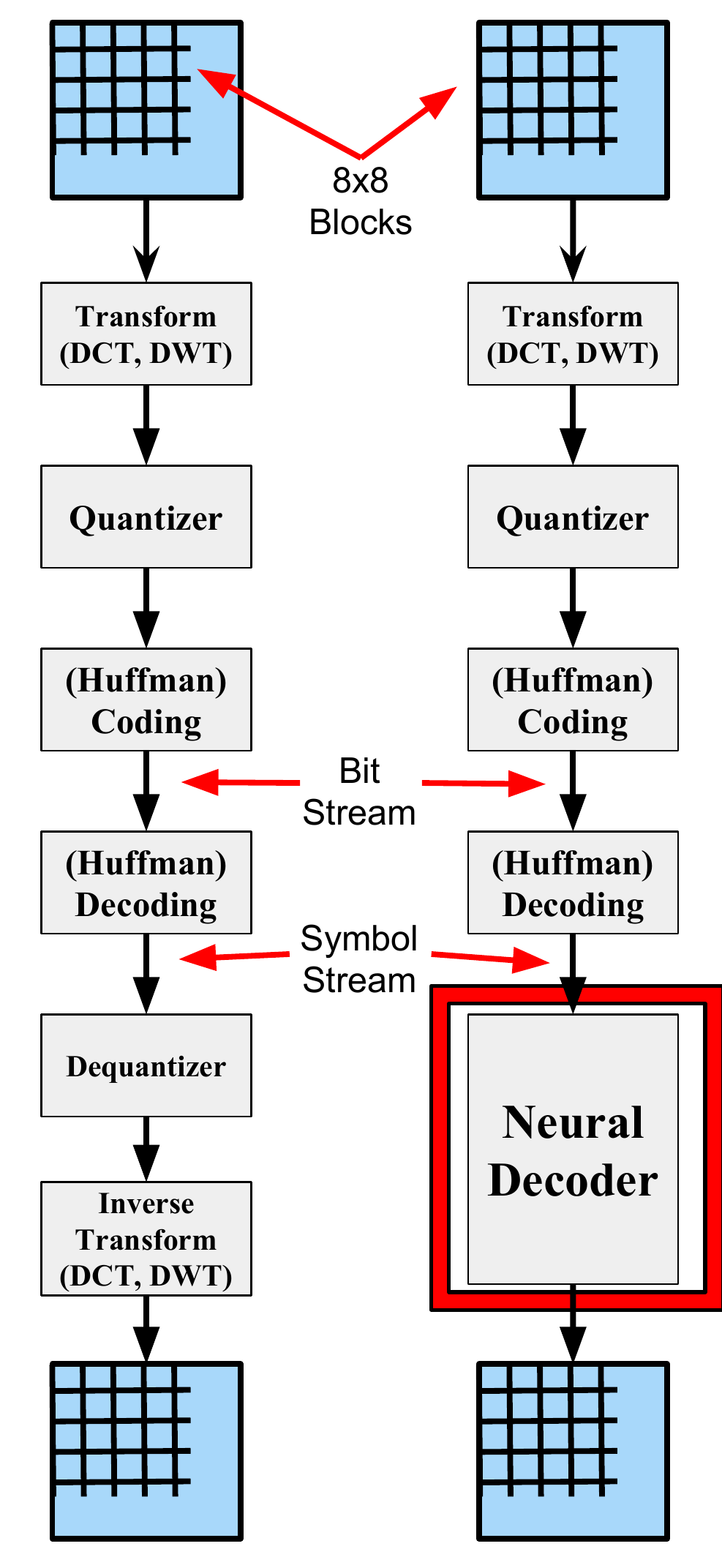}
\includegraphics[width=12.25cm]{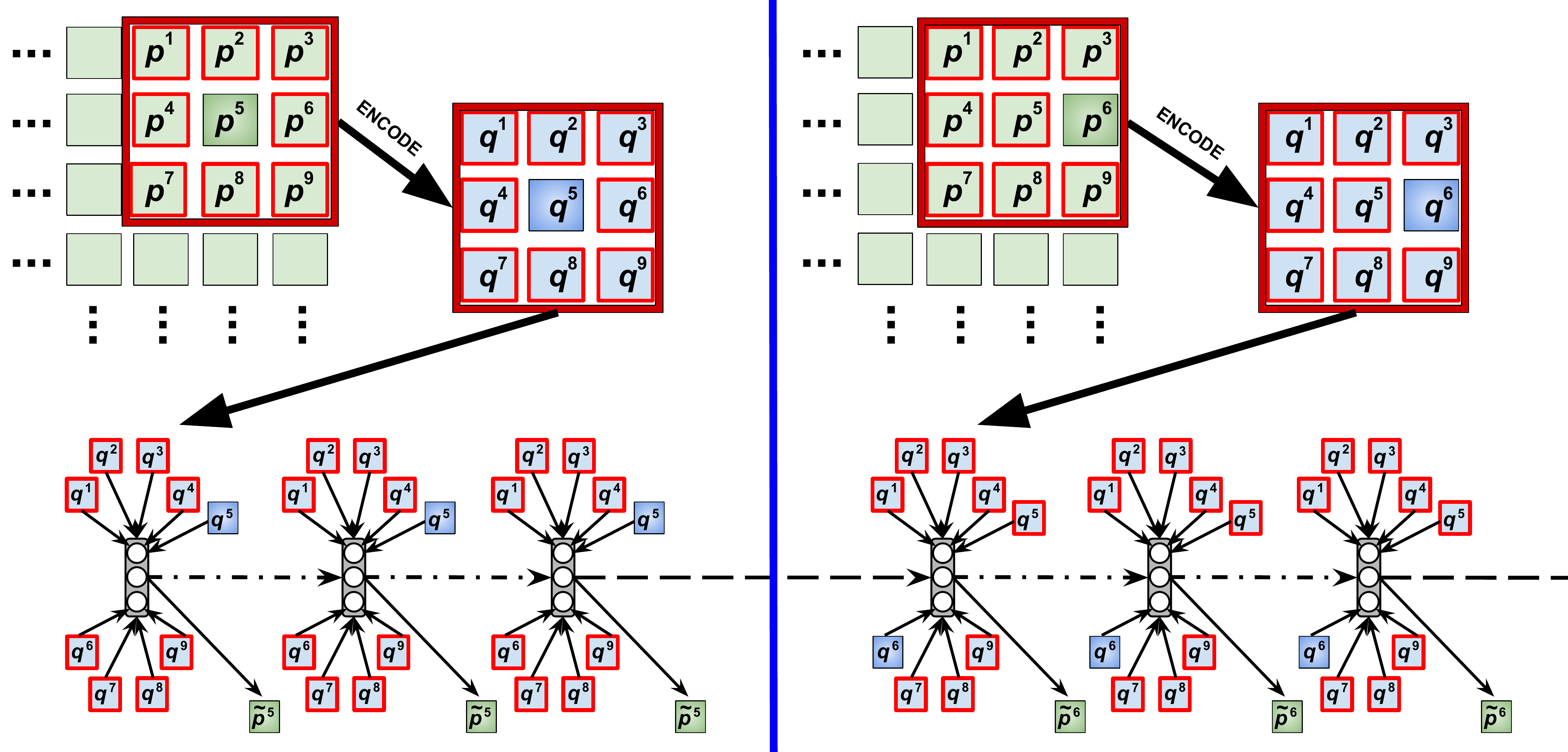}
\caption{Iterative refinement ($2$ episodes, $K=3$) using an Elman RNN estimator (top-level view). 
%Two reconstruction episodes ($K=3$) are depicted. Nine patches are selected, with the patch shaded in green chosen as the reconstruction target. These patches will be fed through an encoder to get a quantized symbol representation that will be presented to the RNN decoder. 
Dot-dashed arrows indicate the latent state across an episode, since back-propagation through time is only applied within an episode. The dashed arrow indicates the memory created at an episode's end, carried over to the next one. This visual also shows how a patch on the borderline is handled.}
\label{fig:sysarch}
\end{figure}

\subsection{The Iterative Refinement Procedure}
\label{refinement}
In order to achieve locally optimal decoding of data points, we propose an iterative, nonlinear process called \emph{iterative refinement}. This process treats the act of reconstructing images from a compressed representation as a multi-step reconstruction problem, which will require a model to improve its recreation of some target sample over a bounded number of passes, $K$. 
%In order to do so, the model will have access to spatial context (non-causal), i.e., neighboring quantized blocks, as well as some memory of its previous state (causal).  % <-- AO: perhaps redundant??

With respect to the data, let us decompose a single image $\mathbf{I}$ into a set of $P$ non-overlapping pixel patches, or $\mathbf{I} = \{ \mathbf{p}^1,\cdots,\mathbf{p}^j,\cdots,\mathbf{p}^P \}$. Assuming column-major orientation, each input patch is of dimension $\mathbf{p}^j \in \mathcal{R}^{d^2 \times 1}$ (a flattened vector of the original $d \times d$ pixel grid). Each patch $\mathbf{p}^j$ would have a corresponding latent representation, or rather, a quantized symbol representation $\mathbf{q}^j$ (as given by the decoding of the bit-stream input to the overall decoder), of dimension $\mathbf{q}^j \in \mathcal{R}^{d^2 \times 1}$. We could furthermore treat these variables $\mathbf{p}^j$ and $\mathbf{q}^j$ as mini-batches of $B$ samples (meaning we are operating on $B$ images in parallel), where a particular instance/sample is indexed by the integer $b$, i.e., $\mathbf{p}^{j,b}$, meaning the dimensions of each would then be $\mathbf{p}^j \in \mathcal{R}^{d^2 \times B}$ and $\mathbf{q}^j \in \mathcal{R}^{d^2 \times B}$, respectively.

Our nonlinear estimator for a locally optimal decoder is defined by parameters $\Theta = \{\Theta_s, \Theta_t, \Theta_d\}$ and takes in $N$ neighboring patches/blocks as input. The overall estimator's general form requires three key functions: 
\begin{itemize}[noitemsep] %[noitemsep,nolistsep]
\item $\mathbf{e} = e(\mathbf{q}^1, \cdots, \mathbf{q}^N ; \Theta)$, a transformation function of a set of encoded neighboring patches (in the form of quantized symbol sequence vectors) for a current target patch that yields a vector summary of the spatial context.
\item $\mathbf{s}_k = s(\mathbf{e}, \mathbf{s}_{k-1} ; \Theta)$, a state function that combines a filtration with the vector summary of input spatial context.
\item $\widetilde{\mathbf{p}}^j_k = d(\mathbf{s}_k ; \Theta)$, a reconstruction function that predicts a given target (pixel) patch at step $k$ (within a reconstruction episode).
\end{itemize}
These ingredients highlight that our estimator requires a way to summarize (encoded) spatial context and state information as well as a transformation of the current state to pixel space. % (generating the prediction $\widetilde{\mathbf{p}}^j_k$ for a target patch $\mathbf{p}^j$).
%%If we take a connectionist perspective to the problem, we may parametrize any or all parts of this decoder as an artificial neural network, we may employ back-propagation of errors to efficiently learn the unrolled computation of multi-step reconstruction.

Using the general specification of the nonlinear decoder above, we can describe the full algorithm for iterative refinement in reconstructing images that are decomposed into multiple, spatially-related patches. First, the encoder, e.g., JPEG, JPEG 2000, is used to obtain an \emph{initial} compressed representation of each patch within the original image, i.e., the bit-stream, followed by decoding, e.g., arithmetic or Huffman, of the bit-stream to a quantized symbol sequence. Then, following some defined scanning pattern, each patch of the image will serve as a target for reconstruction for the decoder, of which a set of neighboring quantized symbol blocks will be fed into the neural estimator. The model then works to best estimate each original image patch. We define the computation for each image patch as a \emph{reconstruction episode} (Figure~\ref{fig:sysarch}). However, when moving on to a new target, the decoder state carries over, i.e., the model is conditioned on a vector summary of its past reconstruction episodes. % bitstream is normally communicated to a decoder when decompressing)

%Why would a stateful decoder prove useful in a static problem such as image decoding?
% Across-episode justification for RNN / connectivity among decoding episodes
It is this carry-over of state across episodes that we conjecture gives a fundamental advantage to an RNN-based estimator over an MLP-based one. When decoding a particular patch, an RNN can exploit global state information, from across the whole image (not just a particular patch's neighborhood), through the iterative propagation of state information from all patches to the patches in their neighborhoods. A stateless MLP, in contrast, can only make use of the information contained in a given patch's surrounding neighborhood. Consider the situation where the model is to decode a particular patch $\mathbf{p}^A$ (having just finished decoding $\mathbf{p}^B$, and before that, patches $\mathbf{p}^D$ and $\mathbf{p}^E$), where $\mathbf{p}^B$ and $\mathbf{p}^C$ are in $\mathbf{p}^A$'s neighborhood but $\mathbf{p}^D$ and $\mathbf{p}^E$ are only in $\mathbf{p}^B$'s neighborhood. The MLP can only use (direct) information from $\mathbf{p}^B$ and $\mathbf{p}^C$ when decoding $\mathbf{p}^A$, treating this reconstruction episode independently of the ones it finished before.
The RNN, however, through the global connectivity provided through its neighborhood function, can also make use of information from $\mathbf{p}^D$ and $\mathbf{p}^E$, since their information was used to update the model's state for $\mathbf{p}^B$ which is used to decode $\mathbf{p}^A$. Note that one could instead modify the stateless MLP to also work with global context, but this would be highly impractical as the model input dimensionality would explode given that we would have to learn a static mapping for each patch position (since the number of patches to the left, right, top, and bottom for a given a patch is position-dependent). The RNN is a practical way to efficiently exploit global context information as opposed to its array of patch position-dependent MLPs equivalent.
% Within-episode justification for the K steps of processing
To justify the use of $K$ steps of recurrent processing within a decoding episode, we can view the $K$ steps as implicitly forming a $K$-layer deep MLP, containing $K$ output predictors, with parameters aggressively shared across its layers (this is a similar idea to the equivalence shown between a shallow RNN and a deep residual network \cite{liao2016bridging}). This equivalence allows us to think of the multiple iterations as simply one compact, nonlinear model (conducting a form of posterior sharpening) that can be learned without the extra memory cost that comes with adding extra layers/parameters. %Furthermore, one could think of the multiple applications of the RNN as a form of posterior sharpening, with the recurrent weight matrix acts like a lateral inhibitory matrix (within an episode).

\begin{algorithm*}[t]
%\scriptsize
\begin{algorithmic}
\State \textbf{Input:} Image represented as $\mathbf{I} = (\mathbf{p}^1, \cdots, \mathbf{p}^j, \cdots, \mathbf{p}^P)$, current decoder parameters $\Theta_{t}$, and \# of steps $K > 0$. Note: $\mathbf{Q} = (\mathbf{q}^1, \cdots, \mathbf{q}^j, \cdots, \mathbf{q}^P)$ is set of quantized symbol blocks. %(after decoding procedure on the bit-stream input). 
%$\widetilde{\mathbf{I}}$ is the output of the algorithm.

%\LineComment method for prediction via iterative refinement
\Function{reconstruct}{$\mathbf{I}$, $\Theta_{t}, K$}
	\State $\mathbf{Q} \leftarrow \Call{GetQuantRep}{\mathbf{I}}$ \Comment Get quantized symbol representation via JPEG/JP2
    \State $\mathbf{s}_0 = 0$ \Comment Initialize state
	\For{$\mathbf{q}^j \in \mathbf{Q}$} \Comment Blocks extracted depend on scan-line path
    	\State $(\mathbf{q}^1, \cdots, \mathbf{q}^N) \leftarrow \Call{getNeighbors}{\mathbf{q}^j, \mathbf{Q}}$
        \LineComment Begin reconstruction episode
        \State $\mathbf{e} = e( \mathbf{q}^1, \cdots, \mathbf{q}^N ; \Theta)$ \Comment Compute pre-activation for spatial context
        \For{$k = 1$ to $K$} \Comment Conduct $K$ steps of refinement
        	\State $\mathbf{s}_k = s(\mathbf{e}, \mathbf{s}_{k-1} ; \Theta)$, $\widetilde{\mathbf{p}}^j_k = d(\mathbf{s}_k ; \Theta)$
        \EndFor
        \State $\mathbf{s}_0 = \mathbf{s}_k$, $\widetilde{\mathbf{p}}^j = \widetilde{\mathbf{p}}^j_K$ \Comment Store final decoder state \& reconstruction
	\EndFor
	\State \Return $(\widetilde{\mathbf{p}}^1, \cdots, \widetilde{\mathbf{p}}^j, \cdots, \widetilde{\mathbf{p}}^P)$ \Comment this output is also denoted as $\widetilde{\mathbf{I}}$
\EndFunction

\end{algorithmic}
\caption{The Iterative Refinement algorithm.}
\label{algo:iter_refine}
\end{algorithm*}

At a high-level, our iterative refinement procedure, depicted in Algorithm~\ref{algo:iter_refine}, takes in as input an image $I$ (or mini-batch of images), current model parameters $\Theta_t$, and a predefined number of steps $K$. First, it creates a quantized symbol representation of the input patches using the subroutine $\Call{getQuantRep}{\cdot}$, which amounts to using the encoder of a given compression algorithm, i.e., contains a transform, a quantizer, and a coding procedure such as turbo coding \cite{mitran2002turbo}, and its corresponding bit-stream decoder. For each target patch $\mathbf{p}^j$, the subroutine $\Call{getNeighbors}{\cdot}$ is called to extract its local context of $N$ contiguous image patches, or rather, their quantized representations, $(\mathbf{q}^1, \cdots, \mathbf{q}^N)$.\footnote{This also returns the quantized symbol representation $\mathbf{q}^j$ of $\mathbf{p}^j$, also fed into our nonlinear estimator.} Note that (if $\mathbf{I} = \mathbf{Q}$, i.e., starting from a compressed image) one could modify the line with $\Call{getQuantRep}{\cdot}$ to instead just call the relevant procedure for bit-stream decoding. % <-- this is decoding-only (starting from something like a JPEG-encoded image/file)

\subsubsection{Transformation \& Reconstruction Function Forms}
Both the transformation function $\mathbf{e} = e(\mathbf{q}^1, \cdots, \mathbf{q}^N ; \Theta)$ and the reconstruction function $\mathbf{s}_k = s(\mathbf{e}, \mathbf{s}_{k-1} ; \Theta)$ can be parametrized by multilayer perceptrons (MLPs). 
%Given constraints on computational resources, we opted for simple versions of each of these functions in favor of more complex state functions, which we describe in the next sub-section.
Specifically, we simply parametrize the transformation and reconstruction functions, using parameters $\Theta_t = \{W_1, \cdots, W_N\}$ and $\Theta_d = \{U, \mathbf{c}\}$, respectively, as follows:
%{\footnotesize
\begin{equation}
\mathbf{e} = \phi_e(W_1 \mathbf{q}_1 + \cdots + W_N \mathbf{q}_N) \mbox{, and, } 
\widetilde{\mathbf{p}}_k = \phi_d(U \mathbf{s}_k + \mathbf{c})
\nonumber %\label{decoder:simple}
\end{equation}
%}
where $\phi_e(v) = v$ and $\phi_d(v) = v$, i.e., identity functions, meaning that nonlinear behavior will come from our design of the state-activation function rather than the input block-to-state or state-to-output mappings. The number of required parameters could be significantly cut down by tying the block-to-hidden matrices, i.e., $W = W_1 = W_2 = \cdots = W_N$.

% AO: condense this down to a single section (use Delta-RNN to unify all models and omit LSTM/GRU equation sets)
\subsubsection{State Function Forms}
\label{state_fun}
%To explore the advantages of increasingly complex state functions, 
We experimented with a variety of gated recurrent architectures as unified under the Differential State Framework (DSF) \cite{ororbia2017diff}. A DSF neural model is generally defined as a composition of an inner function that computes state proposals and an outer mixing function that decides how much of the state proposal is to be incorporated into a slower moving state, i.e., the longer term memory. 
%when faced with the vanishing gradient problem induced by the difficulty of learning long-term dependencies over sequences \cite{bengio1994learning}. 
The DSF models that we will use include the Long Short Term Memory (LSTM) model \cite{hochreiter1997long}, the Gated Recurrent Unit (GRU) \cite{chung2014empirical}, and the Delta-RNN (Delta-RNN or $\Delta$-RNN). We will compare these RNN-based models to a static mapping function learned by a stateless MLP (very much in the spirit of classical predictive coding \cite{spratling2017pred}).
%While the length of each reconstruction episode in our iterative refinement algorithm is not necessarily long enough to create truly far-reaching dependencies, we found experimentally that gated architectures offered the best decoding performance.
%Gated models like the GRU and LSTM were shown to be special cases of the Differential State Framework (DSF) \cite{ororbia2017diff}, which characterized neural models of sequential learning as a composition of an inner function that computes state proposals and an outer mixing function that decides how much of the state proposal is to be incorporated into a slower moving state, i.e., the longer term memory. 
%A vastly simpler and far more parameter-efficient model, the $\Delta$-RNN, was derived from this same framework and shown to outperform complex DSF models like the LSTM and GRU in the task of language modeling (and comparably to intricately regularized variants) \cite{ororbia2017diff} and in text synthesis for protecting user privacy \cite{ororbia2016using}.

We describe the $\Delta$-RNN form, which is the simplest, most parameter-efficient model we experimented with, since models like the LSTM and GRU can be derived from the same framework as that of the $\Delta$-RNN \cite{ororbia2017diff} and integrated into iterative refinement in a similar fashion. The $\Delta$-RNN state function (parameters $\Theta_s = \{V, \mathbf{b}, \mathbf{b}_r, \alpha, \beta_1, \beta_2 \}$) is defined as:
%\footnote{This means a rank-1 matrix approximation of a second-order recurrence \cite{giles1992learning} is used for the $\Delta$-RNN's fast-state.}
%{\footnotesize
\begin{align}
\mathbf{d}^1_k &= \alpha \otimes V \mathbf{s}_{k-1} \otimes \mathbf{e}, \quad 
\mathbf{d}^2_k = \beta_1 \otimes V \mathbf{s}_{k-1} + \beta_2 \otimes \mathbf{e} \\
\widetilde{\mathbf{s}}_k &= \phi_s(\mathbf{d}^1_k + \mathbf{d}^2_k + \mathbf{b}) \\
\mathbf{s}_k &=  \Phi( (1 - \mathbf{r}) \otimes \widetilde{\mathbf{s}}_k + \mathbf{r} \otimes \mathbf{s}_{k-1} ), \ \mbox{where,} \ \mathbf{r} =  \sigma(\mathbf{e} + \mathbf{b}_r) \label{general_second_order},
\end{align}
%}
where $\Phi(v) = \phi_s(v) = tanh(v) = (e^{(2v)} - 1) / (e^{(2v)} + 1)$ and $\otimes$ denotes the Hadamard product.

\subsection{Learning the Neural Iterative Decoder}
\label{sec:learning}
To calculate parameter updates for a particular $K$-step reconstruction episode, one explicitly unrolls the estimator over the length of the episode to create a mini-batch of length $K$ arrays of $B$ matrices (or set of 3D tensors) in order to invoke back-propagation to learn the estimator's parameters, $\Theta$. 
%The cost function, in this case, would follow from that usually posed for any lossy compression system--a linear combination of distortion $D$ and bitrate $R$, regulated by $\lambda$, yielding the unconstrained Lagrangian $L=D+\lambda R$. The basic goal is, for a given $R$, to determine the minimum possible distortion, as measured by mean squared error (MSE). In this paper, however, we do not optimize the bit-rate $R$, meaning that our nonlinear estimator must learn to deal with a variable bit-rate, as dictated by the training sample. 
Our objective will be to optimize distortion $D$, since we are designing a general estimator only for the act of decoding. Note that our estimator must learn to deal with a variable bit-rate, as dictated by training samples.  
The mean bit-rate of our training dataset (described in detail later) was $R_\mu = 0.525$ with variance $R_{\sigma^2} = 0.671$. % with a range of $[0.375-0.875]$.
%Since we are designing a general estimator for decoding only the term $\lambda \mathbf{R}$ effectively disappears since it will be fixed when using any particular encoder.

% (i.e., a meager $0.2$ dB gain in most cases). 
When training decoders using only mean squared error (MSE), we often found that insufficient gains were made with respect to PSNR over JPEG. Motivated to overcome initially disappointing results, we looked to designing a cost function that might help improve our decoders' performance.\footnote{At test time, performance will still be evaluated by strictly measuring MSE \& PSNR.} At training time, we optimize decoder parameters with respect to the multi-objective loss over $K$-step reconstruction episodes for mini-batches of $B$ target image patches $\mathbf{p}^j$ (channel input) operating over a set of decoder reconstructions $\widehat{\mathbf{p}}^j = \{ \widetilde{\mathbf{p}}^j_k, \cdots, \widetilde{\mathbf{p}}^j_K \}$ (channel outputs). We defined the multi-objective loss over a single episode to be a convex combination of MSE and a form of mean absolute error (MAE) as follows:
%{\footnotesize
\begin{align}
\mathcal{D}(\mathbf{p}^j,\widehat{\mathbf{p}}^j) = (1 - \alpha) \mathcal{D}_{MAE}(\mathbf{p}^j,\widehat{\mathbf{p}}^j) + \alpha \mathcal{D}_{MSE}(\mathbf{p}^j,\widehat{\mathbf{p}}^j) \mbox{.} \label{cost_function}
\end{align}
%}
$\alpha$ is a tunable coefficient that controls the trade-off between the two distortion terms. In preliminary experiments, we found that $\alpha = 0.235$ provided a good trade-off between the two (using validation set MSE as a guide).
The individual terms of the cost are explicitly:
%{\footnotesize
\begin{align}
\mathcal{D}_{MAE}(\mathbf{p}^j,\widehat{\mathbf{p}}^j) = \frac{1}{(2 K)}  \sum^K_{k=1} \sum^B_{b=1} \sum_i | ( \widehat{\mathbf{p}}^{j,b}_k[i] - \mathbf{p}^{j,b}[i] ) | \label{eqn:mae} \\
\mathcal{D}_{MSE}(\mathbf{p}^j,\widehat{\mathbf{p}}^j) = \frac{1}{(2 B K)}  \sum^K_{k=1} \sum^B_{b=1} \sum_i ( \widetilde{\mathbf{p}}^{j,b}_k[i] - \mathbf{p}^{j,b}[i] )^2 \mbox{.} \label{eqn:mse}
\end{align}
%}
where $i$ indexes a single dimension of a given vector. 
%%Observe that this term acts like MAE when averaged over the length of an episode but at each step is an L1 penalty.
We speculate that the MAE-based term helps to train a better decoder by reducing outlier errors. MAE is connected to the least absolute deviations statistical optimality criterion, which, in contrast to MSE (least squares), is robust/resistant to outliers in data since it gives equal emphasis to all observed patterns (whereas in least squares, the squaring operator gives more weight to large residual errors). Since PSNR is a function of MSE (and, as is known in regression, using MAE alone can lead to unstable solutions), we felt it unwise to eschew MSE completely and thus combined it with MAE to create a hybrid function.
%However, as verified by our experiments in the appendix, using MAE alone is insufficient, though it does lead to better PSNR than using MSE in training alone. Since PSNR is a function of MSE (and completely ignoring outliers is not always the best move, since MAE does not always lead to stable solutions), we felt it unwise to completely eschew MSE in the training objective, and thus opted for an interpolation between the MAE and MSE metrics with a tunable coefficient that controls the importance of each term in the optimization process.\footnote{We also observed improved convergence when using the combined objective.}
To the best of our knowledge, we are the first to apply such a hybrid objective to the domain of image compression.  
%\footnote{See the appendix for an empirical comparison  of the full proposed cost function versus each term in isolation.} 
%Both metrics are used to compare the channel input with the channel output $\tilde{\mathbf{p}}_k$. 

Learning decoder parameters can be done under an empirical risk minimization framework where we calculate derivatives of our cost with respect to $\Theta$ using back-propagation through time (BPTT) \cite{werbos1988bptt}. Given that the decoder's state function is recursively defined, we unroll the underlying computation graph over a length $K$ reconstruction episode. With the exception of the very first target, we initialize the initial decoder hidden state with its state at the end of the previous episode. The cost of calculating the gradients may be further reduced by pre-computing the linear combination of projected input blocks, i.e., the decoder function $\mathbf{e} = e(\mathbf{q}^1, \cdots, \mathbf{q}^N ; \Theta)$, and reuse $\mathbf{e}$ over each step of the unrolled graph. The BPTT-computed gradients can then be used to update parameters using the method of steepest descent. 
Note that, since $\mathbf{p}^j$ could be a mini-batch, decoding (Algorithm \ref{algo:iter_refine}) and parameter updating can be carried over $B$ images so long as all images are of the same dimensions.\footnote{The patch-by-patch pathway taken across each image within a mini-batch does not have to be the same. One could, in a batch of say 2 images, start from the bottom-left in the 1st and the top-right in the 2nd.% (so long as the pathway is unbroken).
%The only requirement is that the pathway taken must be unbroken in the case of the RNN-based implementations of Algorithm~\ref{algo:iter_refine}.
} 
%In this case, we apply Algorithm~\ref{algo:iter_refine} to each image in parallel, allowing us to create mini-batches $B$ patches.

\section{Experiments}
\label{experiments}
We implement variations of the estimator state-function described above, i.e., a GRU, an LSTM, and a $\Delta$-RNN.
%We design different decoders based on the state functions described above. Specifically, we implemented and trained the GRU, LSTM, and the $\Delta$-RNN variations of the estimator state-function. 
%The first variation $\Delta$-RNN refers to using simple Xavier initialization and $\Delta$-RNN-ortho refers to using orthogonal initialization of the parameters.
We compare these models to the JPEG and JPEG 2000 (JP2) baselines as well as an MLP stateless decoder.  Furthermore, we ran the competitive neural architecture proposed in \cite{toderici2016full} (GOOG) on our test-sets. All of our decoders/estimators (including the MLP) were all trained using the same cost function (Equation \ref{cost_function}). The only pre-processing we applied to patches was normalizing the pixel values to the range $[0,1]$ (for evaluation we convert decoder outputs back to $[0,255]$ before comparing to original patches).

% AO: MARK <-- finish edits here...
\subsection{Data \& Benchmarks}  
\label{exp:data}
To create the training set for learning our nonlinear decoders, we randomly choose high resolution 128k images from the \emph{Places365} {\cite{zhou2017places}} dataset, which were down-sampled to $512\times512$ pixel sizes. In addition, we randomly sampled 7168 raw images with variable bit rates from the RAISE-ALL{\cite{Raise}} dataset, which were down-sampled to a $1600\times1600$ size. Each image was first compressed with variable bit rates between $0.35$--$1.02$ bits per pixel (bpp) for a particular encoder (once for JPEG and once for JPEG 2000). Similarly, to create a validation sample, we randomly selected 20K images from the \emph{Places365} development set combined with the remaining 1K RAISE-ALL images. Validation samples were also compressed using bitrates between $0.35$--$1.02$ bpp. 
For simplicity, we focus this study on single channel images and convert each image to gray-scale. However, though we focus on gray-scale, our proposed iterative refinement can be used with other formats, e.g., RGB. We divide images into sets of $8\times8$ non-overlapping patches, which produces $4096$ patches for images of dimension $512\times512$ and $40000$ for images of size $1600$.

We experimented with $6$ different test sets: 1) the Kodak Lossless True Color Image Suite\footnote{\url{http://r0k.us/graphics/kodak/}} (Kodak) with 24 true color 24-bit uncompressed images, 2) the image compression benchmark (CB~8-Bit\footnote{\url{http://imagecompression.info/}}) with 14 high-resolution 8-bit grayscale uncompressed images downsampled to $1200\times1200$ images, 3) the image compression benchmark (CB~16-Bit) with 16-bit uncompressed images also downsampled to $1200\times1200$, 4) the image compression benchmark 16-bit-linear (CB~16-Bit-Linear) containing 9 high-quality 16-bit uncompressed images downsampled to $1200\times1200$, 5) Tecnick {\cite{teck}} (36 8-bit images), and 6) the Wikipedia test-set created by crawling 100 high-resolution $1200\times1200$ images from the Wikipedia website.

\setlength{\tabcolsep}{4pt}
\begin{table}[t]
\centering\footnotesize
\caption{\footnotesize PSNR of the $\Delta$-RNN-JPEG on the Kodak dataset (bitrate $0.37$ bpp) as a function of $K$. 
%Increasing $K$ allowed per reconstruction episode reduces the distortion of the learned iterative decoder.
}
\label{results:kodak_progressive}
\vspace{-0.425cm}
\begin{tabular}{|c||c|c|c|c|c|c|}
\hline
   & $K = 1$ & $K = 3$ & $K = 5$ & $K = 7$ & $K = 9$ & $K = 11$\\
  \hline
  PSNR &  27.0087&  27.3976 &  27.6619 &  27.8954 &  28.2189 &  28.5093\\
  \hline
\end{tabular}
\vspace{-0.275cm}
%%\end{table}
%%\setlength{\tabcolsep}{1.4pt}

%%\setlength{\tabcolsep}{4pt}
%%\begin{table}[t]
%%\centering\footnotesize
\caption{\footnotesize Out-of-sample results for the Kodak (bpp $0.37$), the 8-bit Compression Benchmark (CB, bpp, $0.341$), the 16-bit and 16-bit-Linear Compression Benchmark (CB) datasets (bpp $0.35$ for both), the Tecnick (bpp $0.475$), and Wikipedia (bpp $0.352$) datasets.}
\label{results:benchmarks}
\resizebox{\columnwidth}{!}{%
\begin{tabular}{|l||c|c|c|c||c|c|c|c|}
\hline
  & \multicolumn{4}{c||}{\textbf{Kodak}} & \multicolumn{4}{c|}{\textbf{CB 8-Bit}}\\
  \textbf{Model} & \textbf{PSNR} & \textbf{MSE} & \textbf{SSIM} &  $MS^3IM$ & \textbf{PSNR} & \textbf{MSE} & \textbf{SSIM} & $MS^3IM$\\
  \hline
  \emph{JPEG} & 27.6540 & 111.604 & 0.7733 & 0.9291 & 27.5481 & 114.3583 & 0.8330 & 0.9383 \\
  \emph{JPEG 2000} & 27.8370 & 106.9986 & 0.8396 & 0.9440 & 27.7965 & 108.0011 & 0.8362 & 0.9471 \\
  \emph{GOOG}-JPEG & 27.9613 & 103.9802 & 0.8017 & 0.9557 & 27.8458 & 106.7805 & 0.8396 & 0.9562 \\
  \emph{MLP}-JPEG & 27.8325 & 107.1089  & 0.8399 & 0.9444 & 27.8089 & 107.6923  & 0.8371 & 0.9475 \\ 
  $\Delta$-\emph{RNN}-JPEG & 28.5093 & 101.9919 & 0.8411 & 0.9487 & 28.0461 & 101.9689 & 0.8403 & 0.9535 \\
  \emph{GRU}-JPEG & 28.5081 & 102.0017 & 0.8400 & 0.9474 & 28.0446 & 102.0041 & 0.8379 & 0.9533 \\
  \emph{LSTM}-JPEG & 28.5247 & 101.9918 & 0.8409 & 0.9486 & 28.0461 & 101.9686 & 0.8371 & 0.9532 \\
  \emph{LSTM}-JP2 & \textbf{28.9321} & \textbf{98.9686} & \textbf{0.8425} & \textbf{0.9496} & \textbf{28.0896} & \textbf{100.9521} & \textbf{0.8389} & \textbf{0.9539} \\ 
  
  \hline
    & \multicolumn{4}{c||}{\textbf{CB 16-Bit}} & \multicolumn{4}{c|}{\textbf{CB 16-Bit-Linear}}\\
  \hline
  \emph{JPEG} &27.5368 & 114.6580 & 0.8331 & 0.9383 & 31.7522 & 43.4366 & 0.8355 & 0.9455 \\
  \emph{JPEG 2000} & 27.7885 & 108.2001 & 0.8391 & 0.9437 & 32.0270 & 40.7729 & 0.8357 & 0.9471 \\
  \emph{GOOG} & 27.8830 & 105.8712 & 0.8391 & 0.9468  & 32.1275 & 39.8412 & 0.8369 & 0.9533 \\
  \emph{MLP}-JPEG & 27.7762 & 108.5056  & 0.8390 & 0.9438 & 32.0269 & 40.7746  & 0.8356 & 0.9454 \\
  $\Delta$-\emph{RNN}-JPEG & 28.0093 & 102.8369 & 0.8399 & 0.9471 & 32.4038 & 37.3847 & 0.8403 & 0.9535 \\
  \emph{GRU}-JPEG & 28.0081 & 102.8649 & 0.8392 & 0.9469 & 32.4038 & 37.3844 & 0.8379 & 0.9533 \\
  \emph{LSTM}-JPEG & 28.0247 & 102.4710 & 0.8310 & 0.9471 & 32.4032 & 37.3908 & 0.8371 & 0.9532 \\
  \emph{LSTM}-JP2 & \textbf{28.1307} & \textbf{100.0021} & \textbf{0.8425} & \textbf{0.9496} & \textbf{32.4998} & \textbf{36.5676} & \textbf{0.8382} & \textbf{0.9541} \\
  
  \hline
  & \multicolumn{4}{c||}{\textbf{Tecnick}} & \multicolumn{4}{c|}{\textbf{Wikipedia}}\\
  \hline
  \emph{JPEG} &  30.7377  & 54.8663 & 0.8682 &  0.9521 & 28.7724  & 86.2655  & 0.8290 &  0.9435\\
  \emph{JPEG 2000} & 31.2319 & 48.9659 & 0.8747 & 0.9569 & 29.1545 & 79.0002 & 0.8382 & 0.9495 \\
  \emph{GOOG} & 31.5030 & 46.0021 & 0.8814 & 0.9608 & 29.2209 &  77.108  &  0.8406 &  0.9520 \\
  \emph{MLP}-JPEG & 31.2287 & 49.0012 & 0.8746 &  0.9571 & 29.1547  & 78.9968 &  0.8383 &  0.9497 \\
  $\Delta$-\emph{RNN}-JPEG &  31.5411  & 45.6001  & 0.8821 &  0.9609 & 29.2772  & 76.8000  & 0.8403  & 0.9519\\
  \emph{LSTM}-JPEG & 31.5616 & 45.3857 & 0.8820 & 0.9609 & 29.2771 & 76.8008 & 0.8403 & 0.9519\\
  \emph{LSTM}-JP2 & \textbf{31.6962} & \textbf{44.0012} & \textbf{0.8834} & \textbf{0.9619} & \textbf{29.3228} & \textbf{75.9969} & \textbf{0.8411} & \textbf{0.9526} \\
  \hline
\end{tabular}
}
\end{table}
\setlength{\tabcolsep}{1.4pt}

\subsection{Experimental Setup}
\label{exp:design}
All RNN estimators contained one layer of 512 hidden units (initial state was null). Parameters were randomly initialized from a uniform distribution, $\sim U(-0.054, 0.054)$ (biases initialized to zero). Gradients were estimated over mini-batches of 256 samples, each of which is $24 \times 24$, or 9 parallel episodes over 9 images, and parameters were updated using RMSprop {\cite{Tieleman2012}} with gradient norms clipped to 7. All models were trained over 200 epochs. We experimented with two learning rate ($\eta$) schedules. The first was simple: start $\eta = 0.002$ and after every $50$ epochs, $\eta$ is decreased by $3$ orders of magnitude. 
The second one was a novel scheme we call the \emph{annealed stochastic learning rate}. In this schedule, with $\eta_0 = 0.002$, at the end of each epoch, we stochastically corrupt the step-size: $\eta_t  = \eta_{t-1} + \mathcal{N}(0,\gamma^{t})$, where $t$ marks the current epoch and $\gamma = 0.000001025$ while the $\eta_t$ is further annealed by a factor of $0.01$ every $50$ epochs. We found that the second schedule helped speed up convergence. Furthermore, we developed a unique two-step data shuffling technique to prevent formation of any spurious correlations between images and patches. In the first step, we randomly shuffle images before the start of an epoch and during the second step (within an epoch), we randomly shuffle the scan pathway starting points. Specifically, per image, we randomly start at the top-left, bottom-left, top-right, or the bottom-right image corner.\footnote{If a scan starts on the image's left side, we proceed horizontally then vertically. If starting on the right side, we proceed left horizontally then vertically.}   %\footnote{Note that care must be taken when constructing mini-batches at the patch-level, since we deviate from many approaches that work at the image-level.Images were divided into $24 \times 24$ non-overlapping patches.}

%2] Stochastic Noise with Annealing:- We Train our model with Initial Learning rate 0.02 and after each epoch, we add noise to our learning rate (lr = previous_lr + (decay_factor ** (i/Epoch)) ,where™ decay_factor = 0.000025.Additionally, Annealing is also applied after 90 epoch with a factor of 0.01
%Adding noise to our learning rate helps our model to converge faster and better.
%Decay_rates were based on experiments.

\subsection{Results}
We evaluate our model on 6 datasets: Kodak True color images, the Image Compression Benchmark (8bit, 16bit, and 16bitlinear), Tecnick and Wikipedia. Every model was evaluated using three metrics, as advocated by \cite{ma2016group}. These included PSNR, structural similarity (SSIM), and multi-scale structural similarity (MS-SSIM \cite{wang2004imagequality}, or $MS^3IM$ as shown in our tables). In addition, we report mean squared error (MSE), though PSNR is a function of it. 
%We observe positive improvements in all metrics over the baseline JPEG and the architecture of \cite{toderici2016full}.

In Table \ref{results:benchmarks}, iterative refinement consistently yields lower distortion as compared to JPEG, JPEG 2000, and GOOG. In terms of PSNR (on Kodak) we achieve nearly a $0.9$ decibel (dB) gain (with \emph{LSTM}-JPEG) over JPEG and a $1.0951$ dB gain (with \emph{LSTM}-JP2) over JPEG 2000. With respect to GOOG, our \emph{LSTM}-JP2 estimator yields a gain of $0.9708$ dB. Furthermore, note that our nonlinear decoders are vastly simpler and faster than the complex end-to-end GOOG model, which faces the additional difficulty of learning an encoder and quantization scheme from scratch. The results, across all independent benchmarks/test-sets, for all metrics (PSNR, SSIM, and MS-SSIM), show that decoders learned with our proposed iterative refinement procedure generate images with lower distortion and higher perceptual quality (as indicated by SSIM and MS-SSIM). Moreover, the recurrent estimators outperform the MLP stateless estimator, indicating that a decoder benefits from exploiting both causal and non-causal information when attempting to reconstruct image patches.

In Table \ref{results:kodak_progressive}, for our best-performing $\Delta$-RNN, we investigate how PSNR varies as a function of $K$, the number of steps taken in an episode. We see that raising $K$ improves image reconstruction with respect to dataset PSNR. We also sampled four random images and plotted their PSNR as a function of $K$.\footnote{Available at \url{https://1drv.ms/b/s!AiNRGVbyH-JUjjVxd7R5Vmy6T2NA}} In general, increasing $K$ improves PSNR, but in some cases, we see a diminishing returns effect and even a peak. This might indicate that integrating a decoding-time early-stopping criterion might yield even better overall PSNR.

%%\begin{table*}[t]\renewcommand{\arraystretch}{1.0}
%%\begin{center}
%%\label{results:samples}
%%\caption{Plot of PSNR as a function of iterative decoding step.}\vspace{1em}
%%\footnotesize
%%\begin{tabular}[t]{cl}
%%\hline
%%\includegraphics[width=0.4\textwidth]{figs/05.png} & \includegraphics[width=0.425\textwidth]{figs/plot5.png} \\
%%\hline
%%\includegraphics[width=0.4\textwidth]{figs/11.png}  & \includegraphics[width=0.425\textwidth]{figs/plot11.png} \\
%%\hline
%%\includegraphics[width=0.4\textwidth]{figs/14.png}  & \includegraphics[width=0.425\textwidth]{figs/plot14.png} \\
%%\hline
%%\includegraphics[width=0.4\textwidth]{figs/17.png}  & \includegraphics[width=0.425\textwidth]{figs/plot17.png} \\
%%\end{tabular}
%%\end{center}
%%\end{table*}

\section{Conclusions}
In this paper, we proposed \emph{iterative refinement}, an algorithm for improving decoder reconstruction for lossy compression systems. The procedure realizes a nonlinear estimator of an iterative decoder via a recurrent neural network that exploits both causal and non-causal statistical dependencies in images. We compared our approach to standard JPEG, JPEG-2000, and a state-of-the-art, end-to-end neural architecture and found that our algorithm performed the best with respect to reconstruction error (at low bit rates). 
%Note that further gains are possible with our algorithm by customizing the design for a target bitrate.
Note that our decoder approach is general, which means any encoder can be used, including that of \cite{toderici2016full}. %This, and studies of our proposed estimator for nonlinear iterative decoding on larger-scale problems, will be the subject of future work. % <-- AO: I cut future work given space constraints...

\Section{References}
\bibliographystyle{IEEEbib}
\bibliography{refs}

\end{document}